\documentclass{article}
\usepackage[utf8]{inputenc}
\usepackage{authblk}
\usepackage{xcolor}
\usepackage{graphicx}

\title{\textbf{Using Argumentation Schemes to Model Legal Reasoning}}

\author{Trevor Bench-Capon and Katie Atkinson\footnote{Presented at European Conference on Argumentation, Rome, 2022}}
\affil{Department of Computer Science, University of Liverpool, Liverpool, UK}

\date{}
\begin{document}

\maketitle

\begin{abstract}
    We present argumentation schemes to model reasoning with legal cases. We provide schemes for each of the three stages that take place after the facts are established: factor ascription, issue resolution and outcome determination. The schemes are illustrated with examples from a specific legal domain, US Trade Secrets law, and the wider applicability of these schemes is discussed.
\end{abstract}


\section{Introduction}


Reasoning with legal cases, especially as conducted in common law jurisdictions such as the UK and USA, is a form of argumentation much studied in Artificial Intelligence and in computational argumentation. The formal procedure within which it is conducted and the extensive documentation which records the argument presented for each side and an assessment of these arguments make it a fruitful area for study. As described in \cite{varsava2018realize}, there may be several types of reasoning involved, including the use of rules, the balancing of factors, analogy and the use of policies to achieve particular purposes. All of these have been modelled in AI and Law, and this work suggests that reasoning with legal cases can been seen as going through a series of stages at which different reasoning styles are appropriate. This view will be elaborated in Section~\ref{stages}.

One way of modelling a reasoning task \cite{prakken2010nature} is to present it as a set of argumentation schemes \cite{walton2008argumentation}. In this paper we will use this method to  articulate the reasoning required at each of the stages. 

Although legal reasoning is worthy of study in itself, we believe that the insights are also applicable to other, less formal, domains where it is necessary to balance reasons for and against particular options to come to a decision. 
While there are some similarities with practical reasoning tasks, such as choosing which restaurant to go to (e.g. \cite{atkinson2013distinctive}) and which car to buy, (e.g. \cite{gordon1997zeno}), we suggest that the methods are more applicable to classification tasks, because the use of binding precedents to determine preferences is not applicable in many domains, whereas the examples used to train a classification system can act like a set of precedents in a settled domain of law where preferences are resolved.
Legal reasoning is indeed used as the model for explaining the output of machine learning systems in \cite{prakken2022top} for several  diverse classification tasks: customer churn, poisonous mushrooms and university admissions.

\section{Stages in Reasoning with Legal Cases} \label{stages}

A legal case will be governed by laws which state what the plaintiff or prosecution must show to establish their claim. The law can be seen as a set of definitions and represented as rules \cite{sergot1986british}. At some point, however, the terms in the rules will be undefined \cite{skalak1992arguments}, and it must be determined whether or not these terms apply in the particular case. This gives rise to a set of what are termed \textit{issues} in AI and Law \cite{aleven}.
These issues must be resolved in favour of either the plaintiff or the defendant, and the reasons for each side are termed \textit{factors} in AI and Law \cite{aleven}. Which reasons apply in a given case will be determined by the \textit{facts}. The facts themselves will be decided on the basis of the \textit{evidence} is presented.

Thus deciding a legal case involves \begin{itemize}
    \item accepting facts on the basis of the evidence
    \item ascribing factors  on the basis of the facts
    \item resolving the issues on the basis of the factors
    \item deciding the outcome of the basis of the issues
\end{itemize}

We will not say more about reasoning with evidence here, since it is not especially legal. In common law jurisdictions, deciding which facts to accept is the responsibility of lay jurors, not legal professionals, and the reasoning as to what to believe is no different from that used in everyday life. A set of argument schemes for reasoning with evidence, based on several common schemes, was presented in \cite{bex2003towards}. We will now say some more about the reasoning involved at each of the other stages.

\subsection{From Issues to Outcome} 

\begin{figure}
\center
\includegraphics[scale=1.4]{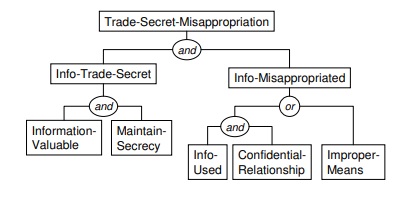} 
\caption{Issues in Us Trade Secrets Law \cite{bruninghaus2003predicting}} \label{IBP}
\end{figure}

Because they are the base level terms in a set of definitions, issues generally can be viewed as supplying necessary and sufficient conditions.  We illustrate this point with reference to US Trade Secrets Law, which has been widely studied in AI and Law (e.g. \cite{rissland1987case}, \cite{ashley1991modeling}, \cite{bruninghaus2003predicting}, \cite{chorley2005empirical}, \cite{prakken2013formalization}, \cite{bench2021precedential}, \cite{prakken2021formal}), and which we will use for our examples in this paper. Cases falling within this domain cover scenarios where there is a claim that a trade secret has been misappropriated, thus the information being considered in a case must be both a \textit{Trade Secret} and have been \textit{misappropriated}. These terms may be further defined: for information to be a Trade Secret it must be both \textit{valuable} and its \textit{secrecy maintained}. For information to be misappropriated it must have been \textit{used} despite a \textit{confidential relationship}, or obtained through \textit{improper means}. Thus the issues can be expressed as what is termed a ``logical model'' in \cite{bruninghaus2003predicting}, as shown in Figure~\ref{IBP}.

Once each of the leaf issues is resolved, the outcome can be determined.

\subsection{From Factors to Issues}

The resolution of the issues, however, tends to be less clear cut. Typically there will be reasons both why the issue should be resolved one way and why it should be resolved the other way. For example, with respect to the existence of a \textit{confidential relationship}, a person may have disclosed the information in negotiations, which is a reason to find for the defendant, but the defendant may have been made aware that the disclosure was in confidence, which is a reason to find for the plaintiff, even if there was no formal agreement. Each issue will thus be associated with a set of factors, providing reasons to resolve the issue either for the plaintiff or the defendant. The factors associated with the issues in US Trade Secrets in \cite{ashley2009automatically} are shown in Table~\ref{IBPIssues}

\begin{table*} [t]
\caption{CATO factors grouped by Issues \cite{ashley2009automatically} \label{IBPIssues}}
\tiny
\begin{tabular}{|l|l|l|}
\hline
Issue                    & Plaintiff Factors                                                                                                                                          & Defendant Factors                                                                                                                                                                   \\ \hline
InfoValuable             & \begin{tabular}[c]{@{}l@{}}F8p Competitive Advantage\\    F15p Unique Product\end{tabular}                                                                 & \begin{tabular}[c]{@{}l@{}}F16d Info Reverse Engineerable\\    F20d Info Known to Competitors\\    F24d Info Obtainable Elsewhere\\    F27d Disclosure In Public Forum\end{tabular} \\ \hline
SecrecyMaintained        & \begin{tabular}[c]{@{}l@{}}F4p Agreed Not To Disclose\\    F6p Security Measures \\F12p Outsider Disclosures Restricted\end{tabular}                                                              & \begin{tabular}[c]{@{}l@{}}F10d Secrets Disclosed Outsiders\\       F19d No Security Measures\end{tabular}                                   \\ \hline
ImproperMeans            & \begin{tabular}[c]{@{}l@{}}F2p Bribe Employee\\  F7p Brought Tools \\  F14p Restricted Materials Used\\    F22p Invasive Techniques\\    F26p Deception\end{tabular}           & \begin{tabular}[c]{@{}l@{}}F17d Info Independently Generated\\    F25d Info Reverse Engineered\end{tabular}                                                                         \\ \hline
InfoUsed                 & \begin{tabular}[c]{@{}l@{}}F7p  Brought Tools\\    F8p Competitive Advantage\\    F14p Restricted Materials Used\\    F18p Identical Products\end{tabular} & \begin{tabular}[c]{@{}l@{}}F17d Info Independently Generated\\    F25d Info Reverse Engineered\end{tabular}                                                                         \\ \hline
ConfidentialRelationship & \begin{tabular}[c]{@{}l@{}}F4p Agreed Not To Disclose\\    F13p Noncompetition Agreement\\    F21p Knew Info Confidential\end{tabular}                     & \begin{tabular}[c]{@{}l@{}}F1d Disclosure In Negotiations\\    F23d Waiver of Confidentiality\end{tabular}                                                                          \\ \hline
\end{tabular}
\end{table*}

These \textit{factors} must be balanced against one another. Which set of reasons is considered stronger will depend on the preferences of the person making the choice: some will require a formal agreement, while others will not. In law these preferences are revealed in the past decisions, and courts are bound by these precedents\footnote{For example, National Instrument Labs, Inc. v. Hycel, Inc., 478 F.Supp. 1179 (D.Del.1979) provides a precedent in favour of the existence of a confidential relationship where there had been \textit{DisclosureInNegotiations} (F1d), but the defendant \textit{KnewInfoConfidential} (F21p).}. Where no precedent exists, the judges must express a preference, creating a precedent that will govern future cases. Many questions arise concerning the construction of preferences in general \cite{lichtenstein2006construction}, but in AI and Law the ideas of \cite{BH93}, \cite{bench2003model} and \cite{grabmair2016modeling} have generally been followed and it is taken that the preference will be according the purpose or social value promoted by the decision. This may involve argument as to which value should be preferred: this is modelled in \cite{bench2009case}. We will not, however, discuss the question further in this paper: our discussion will be in terms of settled law, where precedents are available to resolve such preference questions.

The reasoning at this stage thus involves, for each issue, identifying the relevant factors for and against, and then, in the case of factors for both sides, choosing the stronger set, in accordance with precedents. Formal accounts of this style of reasoning with precedents are given in \cite{horty2012factor} and \cite{prakken2021formal}, and a set of argument schemes modelling it is given in \cite{prakken2013formalization}. These formal accounts are at the whole case level, but can readily be adapted to consider issues rather than complete cases \cite{bench2021precedential}.

\subsection{From Facts To Factors}

What factors are present in a given situation will depend on the facts. But the presence of a factor is not straightforwardly determined by the facts. For example, the security measures taken by the plaintiff will be relevant, but whether they are considered \textit{adequate}, and so a reason to find that secrecy had been maintained and that factor F6p applies, or \textit{inadequate}, and so F19d, a reason to decide that it had not, applies, will depend on what is considered reasonable, and the strictness of the standard applied. Thus the various relevant facts must be assessed for their significance for the decision maker, and this significance recorded by ascribing the appropriate factor to the case. Where the line is drawn will require a judgement. In common law, precedents will determine these ``switching points'' \cite{rigoni2015improved}. 

The ascription of factors has received rather less attention in AI and Law than the other stages. In \cite{bench2022} four different  ways of ascribing factors were described: these will be discussed below in Section~\ref{ascription}

\subsection{Summary}

Different AI and Law approaches have covered different stages in this process, as shown in Table~\ref{layers}. Thus \cite{sergot1986british} models only the legislation (of the British Nationality Act), and does not represent the interpretation of terms in the legislation through case law. In HYPO the dimensions are neutral and the user must decided which party they favour. In CATO the party favoured on a dimension is identified through the use of factors, but the user must decide which side is favoured on the balance of these factors. IBP can predict an outcome by using a logical model of issues once the issues have been resolved on the balance of factors. The focus of Bex \textit{et al.'s} hybrid apprroach \cite{bex2010hybrid} is on the move from evidence to facts.
\begin{table*}[h]
\center
\caption{Layers of Statements in a Legal Decision and Some Example Systems \smallskip} \label{layers}
\begin{tabular}{|l|l|l|l|l|l|}
\hline
Statement Type & BNA \cite{sergot1986british} & HYPO \cite{rissland1987case} & CATO \cite{aleven} & IBP \cite{bruninghaus2003predicting}& Bex \cite{bex2010hybrid} \\ \hline
Outcome        & X   &      &      & X   &              \\ \hline
Issues          & X   &      & X    & X   &            \\ \hline
Factors        &     & X    & X    & X   &               \\ \hline
Facts          &     & X    &      &     & X          \\ \hline
Evidence       &     &      &      &     & X          \\ \hline
\end{tabular}
\end{table*}

In this paper we will model the three stages after  the facts have been agreed, using argumentation schemes and their critical questions \cite{walton2008argumentation}.


\section{Using Argumentation Schemes to Model Reasoning} \label{using}

Argument schemes are typically seen as a form of defeasible inference rules, but as argued in \cite{prakken2010nature} they can also be used to model a reasoning process, by articulating the arguments that can be used and the objections that can be made at each stage of the process. Such sets of argument schemes can be used as the specification of dialogue system to realise this process (e.g. \cite{atkinson2005dialogue}).  Here we will use argument schemes to describe the process of reasoning with legal cases.


To present reasoning with legal cases as an argument, a form of three ply argumentation was introduced in HYPO \cite{rissland1987case}. First a claim is made by a proponent, then the opponent challenges the claim, and finally the proponent tries to rebut these challenges. A repertoire of moves for each of these stages was developed \cite{aleven}. This structure fits well with the notion of argumentation schemes as proposed by Walton \cite{walton2013argumentation}. Here an instantiated scheme is put forward, challenges made in the form of critical questions, and then the proponent attempts to provide answers to these questions. This correspondence allows the argument moves of the original systems to be seen as argumentation schemes. Argumentation schemes provide an excellent way of making a reasoning procedure more precise, and indicating the ways in which assertions within that procedure may be challenged.

In this paper we will describe the three stages of the reasoning with legal cases identified above as argumentation schemes.  


\section{Determining the Outcome}

Once resolved, issues can form the leaves of a logical model as depicted in Figure~\ref{IBP}. This can then be seen as a standard example of logical reasoning, using the law represented as a set of rules. We can express the schemes thus:

\smallskip

\indent
\textbf{Issue to Outcome Scheme (IO)}:

\textit{Warrant Premise}: The law yields a rule $R$: $I_1, I_2 ... I_n \rightarrow O$

\textit{Issue Premise}:  $I_1, I_2 ... I_n$ are (not) satisfied

\textit{Conclusion}: Outcome is (not) $O$

\noindent

\smallskip

Examples  in  US Trade Secrets are ;

\smallskip

\indent

\textbf{Example1}

The law yields a rule that if the information is Trade Secret and Misappropriated the plaintiff should win

The information is Trade Secret and Misappropriated

Therefore, The plaintiff should win

\smallskip
\textbf{Example2}

The law yields a rule that if the information was Used and There is a Confidential Relationship and the information was  Misappropriated the plaintiff should win

The information was not Misappropriated

Therefore, The defendant should win

\noindent

\smallskip
Because these are strict rules, critical questions must take the form of questioning the premises.

\smallskip
\indent
\textbf{IOCQ1}: \textit{Exception}: Is there an issue $I_{n+1}$ which is satisfied and which provides an exception to $R$? (Where the argument is for $O$)

\textbf{IOCQ2}: \textit{Unneeded Premise}: Is there an issue $I_n$ :which is not required for $O$? (Where the argument is for not $O$)

\textbf{IOCQ3}: \textit{Issue Incorrectly Resolved}: Is there an issue $I_n$ which is in fact (not) satisfied?

\noindent

\smallskip

\indent

We could thus pose the following critical questions in Example1:

\smallskip

\textbf{IOCQ1Ex1}: That the employee was the sole developer of the information provides an exception to the rule.

\textbf{IOCQ3Ex1}: The information was not, in fact, misappropriated.

\noindent

\smallskip

and we can object to Example2 with IOCQ2:

\smallskip

\indent

\textbf{IOCQ2Ex2}: It is not necessary that the information be \textit{Misappropriated} where the information was \textit{Used} and there is a \textit{Confidential Relationship}.

\noindent

\smallskip

\section{Resolving the Issues}

As described in \cite{levi1948introduction}, after an initial period of flux sufficient precedents are established for an area of law to be considered settled, until some events bring about a period of reinterpretation. Thus for much of the time, precedents will be available to decide questions of preference between factors: except for newly enacted legislation, landmark cases setting new precedents are relatively rare.


\subsection{Resolving Issues With Precedents}

A set of schemes to balance factors within a case in accordance with a set of precedents  was proposed in \cite{prakken2013formalization}. These, however, ignored the intermediate stage of issues, and moved straight from factors to decisions.
The schemes can, however, be easily adapted to resolve issues by considering not the complete set of factors but only those pertaining to the issue under consideration. It is generally straightforward to allocate factors to issues, as shown in Table~\ref{IBPIssues}. 
Thus in the following $F_p$ and $F_d$ should be taken to comprise only factors relevant to the issue $I$.

\indent

\smallskip

\textbf{Citation Scheme (C):}
\textit{Factor Premise}: Case $C$ has plaintiff factors $F_p$ and defendant factors $F_d$ in common with precedent $P$

\textit{Precedent Premise}: Issue $I$ was resolved for the plaintiff (defendant) in $P$

\textit{Conclusion}: Issue $I$  should be resolved for the plaintiff (defendant) in $C$

\smallskip
\noindent

For an example from US Trade Secrets we will consider resolution of the issue of \textit{Secerecy Maintained}.

\smallskip

\indent

\textbf{Example3}

Among the factors associated with Secrecy Maintained are  the plaintiff factors  \textit{SecurityMeasures} (F6p) and  \textit{OutsiderDisclosuresRestricted} (F12p) and the defendant factor 
\textit{DisclosuresToOutsiders} (F10d). Suppose we have a case, \textit{Restricted}, in which all three of these factors are present. We have a precedent, \textit{Bryce}\footnote{M. Bryce \& Associates, Inc. v. Gladstone, 107 Wis.2d 241, 319 N.W.2d 907 (Wis.App.1982)}, in which \textit{SecurityMeasures} was present and the other two factors were absent, and the issue was resolved for the plaintiff.  On the basis of \textit{Bryce} we can argue:

\noindent

\smallskip

\textit{Restricted} has factor \textit{SecurityMeasures} in common with \textit{Bryce}.

The issue \textit{Secrecy Maintained} was resolved for the plaintiff in \textit{Bryce}.

Therefore, the issue \textit{Secrecy Maintained} was resolved for the plaintiff in \textit{Restricted}. 

\smallskip

\noindent

Two critical questions may be posed against this scheme, either pointing to a counterexample, a precedent in which the issue was resolved differently, or by pointing to another factor present only in the case or the precedent which weakens the argument.

\indent

\smallskip

\textbf{CCQ1}: \textit{Counterexample}: Is there another precedent $P'$ with factors in common with $C$ in which the issues was resolved for the defendant (plaintiff)?

\textbf{CCQ2}: \textit{Distinction}: Is there a factor present in only the case or the precedent which weakens the case for the plaintiff (defendant)?

\smallskip

\noindent

These two critical questions are found in \cite{prakken2013formalization}. But that paper takes the factors present in a case as given. If the presence of a factor can be disputed we get two more critical questions:

\indent

\smallskip

\textbf{CCQ3}: \textit{Factor Not Present}: Is one of the factors $f$ not in fact present in the case?

\smallskip

\noindent

If the conclusion of the critiqued argument is to resolve the issue for the plaintiff, $f \in F_p$, and if to resolve for the defendant, $f \in F_d$.

\indent

\smallskip

\textbf{CCQ4}: \textit{Additional Factor}: Is there an additional weakening factor $f'$  present in the new case?

\smallskip

\noindent

If the conclusion of the critiqued argument if to resolve the issue for the plaintiff, $f'$ favours the defendant, and if to resolve for the defendant, $f$ favours the plaintiff.

\smallskip

Counterexamples to pose CCQ1 are put forward using the Citation Scheme. Distinctions to pose CCQ2 are put forward using their own argument scheme.

\indent

\smallskip

\textbf{Distinction Scheme (D)}:

\textit{Factor Premise A}: There is a factor $F$ present in only one of the current case $C$ and Precedent case $P$

\textit{Polarity Premise:} $F$ weakens the case for the plaintiff (defendant)

\textit{Conclusion}: Do not resolve Issue $I$ for the plaintiff(defendant)

\smallskip

\noindent

The polarity premise is needed because it is possible that the difference will strengthen rather than weaken a case. What is required (when attacking an argument for the plaintiff) is a pro-defendant factor in the current case but not the precedent, or a pro-plaintiff factor in the precedent but not the current case. 

\indent

\smallskip

\textbf{Example3 continued}

\textit{Restricted} has an additional defendant factor, \textit{DisclosuresToOutsiders}, and this can be used to distinguish \textit{Bryce}: 

\smallskip

\textit{DisclosuresToOutsiders} is present in \textit{Restricted} but not in \textit{Bryce}

\textit{DisclosuresToOutsiders} weakens the case for the plaintiff

Therefore, Do not resolve Issue $I$ for the plaintiff.

\smallskip

\noindent

Critical questions can, of course, now be posed against the distinction scheme. These are based on the notion of downplaying a distinction developed in \cite{aleven}. The idea is there may be some other factor present in only one of the cases which strengthens the case for the original party. Such a factor may be used to substitute for a missing factor (if the party favoured is the same), or cancel out an additional factor (if the party favoured is different).

\indent

\smallskip

\textbf{DCQ1}: \textit{Substitution}: Is there a Factor $F'$ which can substitute for $F$?

\textbf{DCQ2};  \textit{Cancellation}: Is there a Factor $F'$ which can be used to cancel $F$?

\smallskip

\textbf{Example3 continued}

In the case of \textit{Restricted} we do have a factor which can cancel out the additional factor 
\textit{DisclosuresToOutsiders}, namely \textit{OutsiderDisclosuresRestricted}, so we can pose DCQ2.

Whether the factor is considered sufficiently strong to cancel out the distinguishing factor is something that must be decided by the court. If it were considered that it was strong enough, it would still be possible to pose CCQ3 by asking whether the security measures taken in \textit{Restricted} were indeed adequate.

\smallskip

\noindent

\subsection{Where Precedents are not Needed}

The above schemes do rely on the existence of precedents to determine the preference for the factors associated with one side rather than the other. Because, as can be seen from Table~\ref{IBPIssues}, there is only a relatively small number of factors associated with each issue, only a few leading cases will be needed to supply the necessary preferences. It should, however, be noted that in some cases the preference will be obvious from the nature of the factors; for example if the plaintiff has given a \textit{WaiverOfConfindentiality} (F23d), then the issue will be resolved for the defendant, even if there had been an agreement not to disclose (F4p), or even a formal non-disclosure agreement (F13p). Such factors were termed \textit{knockout} factors in \cite{bruninghaus2003predicting}. 

For such factors we can have another scheme:

\indent

\smallskip

\textbf{Knockout Factor Scheme (KO)} 

\smallskip

\textit{Factor Premise}: Factor $F$, relating to issue $I$, is present in Case $C$

\textit{Knockout Premise}: Factor $F$ favours plaintiff (defendant) and is, by its nature, preferred to any factors favouring the defendant (plaintiff)

\textit{Conclusion}: Issue $I$ should be found for the plaintiff (defendant)

\smallskip

\noindent

Critical questions can concern either whether the factor is present (to be resolved using the schemes for factor ascription presented in the next section), or whether it is indeed decisive (by citing a counterexample).

\indent

\smallskip

\textbf{KOCQ1}: Is factor $F$ really present in $C$?

\textbf{KOCQ2}: Is there a precedent case $P$ containing factor $F$ in which issue $I$ was resolved for the defendant (plaintiff)?

\smallskip

\noindent

\subsection{Cases which Require a Preference} \label{pref}

If we have an issue which cannot be resolved using either a precedent or a knockout factor, then the judges must themselves express a preference between the pro-plaintiff and pro-defendant factors. As discussed in \cite{BH93} and \cite{bench2003model}, this will involve expressing a preference for the values promoted by finding for the plaintiff or those promoted by finding for the defendant. Such arguments are likely to be very varied, including  such things as an appeal to established values in society (``life is more important than property''), feasibility (``where this to be decided, the floodgates for acrimonious litigation would be opened''), analogy with a different area of law (``this preference is established in contract law and should apply here also'') or the constitutional remit of judges (``such a decision can only be made by the legislature''). The computational deployment of such arguments was discussed in \cite{bench2009case}. The schemes used for such arguments are, however, not specifically legal, and may use a variety of the  established schemes to be found in \cite{walton2013argumentation}. We will therefore not discuss them further here, but limit our discussion to areas of law sufficiently settled that the required precedents are available.

\section {Ascribing the Factors} \label{ascription}

\subsection{Dimensions and Factors}

We now come to the ascription of factors. Factors were a development from dimensions in HYPO. In HYPO dimensions were aspects of cases which were potentially relevant to the outcome. In general, dimensions were a range of values which increasingly favoured one of the parties. Thus the number of disclosures was one such dimension, and the more disclosures, the more favourable the dimension was to the defendant. A small number of disclosures might be considered not to constitute a reason to find for the defendant, but at some point the number of disclosures would be sufficient. If a particular case falls in the range where the defendant is favoured, this means that the pro-defendant factor \textit{DisclosuresToOutsiders} applies, but fewer disclosures means that no factor applies on this dimension. The point at which the factor starts to apply was termed the switching point in \cite{rigoni2018representing}. Most dimensions are either neutral or favour a particular side. Some, however, such as \textit{SecurityMeasures}, favour the plaintiff at one end and the defendant at the other, giving rise to both a pro-plaintiff and a pro-defendant factors, possibly with a neutral range between them. Other dimensions, such as disclosures  give rise to two factor for the same side: if the disclosure is in the public domain the stronger pro-defendant factor  \textit{DisclosureInPublicForum} applies rather than the normal \textit{DisclosuresToOutsiders}.

However, in HYPO many of the dimensions (10 out of the 13)  were, in fact Boolean. Here one of the values would give rise to a factor: thus it is either true or false that the information was disclosed to the defendant in the course of negotiations, and if true then the pro-defendant factor \textit{DisclosureInNegotiations} applies.

\begin{figure}
\center
\includegraphics[scale=0.7]{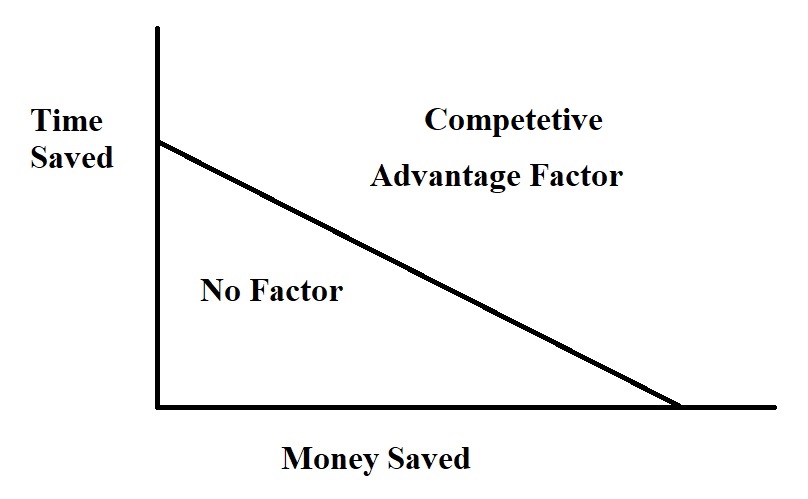} 
\caption{Competitive Advantage Factor} \label{tradeoff}
\end{figure}

Finally we have a factor which arises from two dimensions: use of a trade secret may save the defendant time, money or both. If these savings are significant, then the factor pro-plaintiff \textit{CompetiveAdvantage} (F8p) will apply. The significance will require consideration of both time and money: the more time saved, the less money need be saved and vice versa. Thus we get a trade off of the sort shown in Figure~\ref{tradeoff}

These different relationships between factors and their dimensions mean that there are several ways in which factors are ascribed. 

\begin{itemize}
    \item For Boolean factors, factors may be ascribed on the basis of facts supplying necessary and sufficient conditions (i.e. the factor can be interpreted using its ordinary meaning);
    \item Also Boolean factors which may be  ascribed according to analogy (for example employing a former employee of a competitor at well over the market rate in order to obtain information acquired in the former employment may be considered analogous to bribery).
    \item Factors ascribed according to whether some threshold (switching point) is passed (as in the case of disclosures);
    \item Factors which involve a trade off between two facts (such as time saved  and money saved);
\end{itemize}

Each of these will have their own associated argument scheme.

\subsection{Ordinary Meaning Scheme}

The most straightforward scheme is where the facts of the case justify the ascription of a factor, on a ordinary interpretation of the terms involved.
 \smallskip
 
 \indent 
\textbf{Ordinary Meaning Scheme}

\textit{Facts Premise}: Facts $a_1 ... a_n$ are true in Case $C_1$

\textit{Usage Premise}: As $F$ is ordinarily understood, $a_1 ... a_n$ are sufficient for
Factor $F$ to be considered present in $C_1$

\textit{Conclusion}: $F$ is present in $C_1$
 \noindent
 \smallskip

The following is a set of critical questions to enable the scheme’s components to be questioned:
 
\indent
\textbf{MCQ1}: Does $a_1 ... a_n$ really justify the ascription of $F$? There might be some additional fact which is needed. For example we might require specific mention of the allegedly misappropriated information in a 
non-disclosure agreement.

\textbf{MCQ2}: Does some other fact, $b$, provide an exception which prevents the ascription of $F$? There might be some unusual feature in the situation which should prevent ascription. For example, although the information was disclosed in negotiations, the defendant entered the negotiations under false pretences.

\textbf{MCQ3}: Do other facts $b_1 ... b_n$ justify the ascription of factor $F_2$, which is incompatible with $F$? For example if the defendant had used restricted materials when developing his project, that should not be considered an example of reverse engineering, and so \textit{ResatrictedMaterialsUsed} (F14p) should apply and \textit{ReverseEngineered} (F25d) should not.

\noindent

\subsection{Analogy Scheme}

There are a number of schemes for analogy in the literature. Different schemes are given in \cite{walton2008argumentation}, \cite{walton2010similarity} and \cite{stevens2018reasoning}. Here we give one tailored to our need to analogise between aspects of cases rather than cases as a whole.

\smallskip
\indent
\textbf{Analogy Scheme}

\textit{Base Premise}: A situation $S_1$ is described in precedent $P_1$.

\textit{Derived Premise}: Factor $F$ is plausibly ascribed to $P_1$ on the basis of $S_1$.

\textit{Case Premise}: Case $C_1$ contains situation $S_2$

\textit{Similarity Premise}: As it relates to $F$, situation $S_2$ is similar to situation $S_1$.

\textit{Conclusion}: Factor $F$ is plausibly ascribed to $C_1$.
 \noindent
 \smallskip
 
 \smallskip
 
 \indent
 
 \textbf{Example4}
 
 In \textit{Space Aero}\footnote{Space Aero Products Co. v. R.E. Darling Co., 238 Md. 93, 208 A.2d 74 (1965).}, the defendant had acquired the information while an employee of the plaintiff and the issue of Confidential Relationship was resolved for the plaintiff, on the basis of the factor \textit{KnewInfoConfdential} (F21p). Suppose in a new case, \textit{Subcontract}, the defendant had acquired the information while an employee of a subcontractor working for the plaintiff.
 
 We can now suggest an analogy between the two cases:
 
 \smallskip
 
In \textit{Space Aero}, the information was acquired while an employee of the plaintiff

\textit{KnewInfoConfdential} was ascribed to \textit{Space Aero} on this basis

As it relates to \textit{KnewInfoConfdential}, being an employee of a subcontractor of the plaintiff  is similar to being an employee of the plaintiff

\textit{KnewInfoConfdential} is plausibly ascribed to \textit{Subcontract}

\smallskip

\noindent

 The following set of critical questions is based on the account given in \cite{walton2008argumentation} for the basic scheme for argument from analogy.
 
\indent
\textbf{ACQ1}: Are there respects in which $P_1$ and $C_1$ are different that would tend to undermine the
 force of the similarity with respect to $F$? For example, a sub-contractor has a transient relationship, whereas an employee is in a more stable relationship. 
 
 \textbf{ACQ2}: Is the similarity sufficient for $F$ to be ascribed? Employees of a sub-contractor have no direct relationship to the owner of the information.
 
\textbf{ACQ3}: Is there some other precedent $P_2$ that is also similar to $C_1$, but in which $F$ was not ascribed?  Suppose there was a precedent with a fixed term employee, where the relationship was not considered sufficient to ascribe \textit{KnewInfoConfidential}.
\noindent

\subsection{Switching Point Scheme}
 
 The next scheme is based on Rigoni's notion of a \textit{switching point} \cite{rigoni2018representing}. If we consider a dimension with a factor favouring the plaintiff at one end and a factor favouring the defendant at the other, there will be points (possibly the same) at which one factor ceases to apply and the other factor begins to apply. These are the \textit{switching points}. Thus given a precedent more favourable on the dimension than the new case, we can say that the factor applies to the new case. Similarly, if the new case is less favourable, we can argue that the factor does not apply. We can use this notion as the basis of an argumentation scheme:

\indent

\textbf{Switching Point Scheme}

\textit{Precedent Premise}: $P_1$ is a precedent with location $L_1$ on dimension $D$ at which factor $F$ is present.

\textit{Case Premise}: $C_1$ is a case with $L_2$ on dimension $D$

\textit{Party Premise}: $F$ favours the plaintiff (defendant)

\textit{Value Premise}: $L_2$ is more (less) favourable to the plaintiff (defendant) than $L_1$

\textit{Conclusion}: $F$ applies (does not apply) to $C_1$
\noindent

\smallskip

\indent

\smallskip

\textbf{Example5}

In \textit{National Rejectors}\footnote{National Rejectors, Inc., v. Trieman, 409 S.W.2d 1 (Mo.1966).}
engineering drawings were sent to customers and prospective bidders without limitations on their use. There were perhaps 100 such recipients and it was held that the factor \textit{DisclosuresToOutsiders} applied. Suppose we have a new case, \textit{Leaky}, in which there had been a similar practice with 150 recipients. We can say for the argument:

\smallskip

\textit{National Rejectors} is a case with 100 disclosures and \textit{DisclosuresToOutsiders} is present

\textit{Leaky} is a case with 150 disclosures

\textit{DisclosuresToOutsiders} favours the defendant

150 is more favourable to the defendant than 100

\textit{DisclosuresToOutsiders} applies to \textit{Leaky} 

\smallskip

\noindent

We can question an instantiation of this scheme with the following critical questions:

\indent
\textbf{SCQ1}: \textit{Is $L_2$ so much more favorable that a different factor applies?} For example in the US Trade Secret domain of \cite{aleven} there are two pro-defendant factors on the disclosures dimension, \textit{DisclosedToOutsiders} and the stronger \textit{DisclosedInPublicForum}.

\textbf{SCQ2}: When arguing that the factor does not apply because $L_2$ is less favourable: \textit{Is $L_2$ sufficiently close to $L_1$ that the same factor applies?} It is possible that $P_1$ does not precisely identify the switching point, and that $C_1$ may become a new precedent for the factor, giving a more generous switching point. For example, had there been only 90 disclosures in a new case \textit{LessLeaky}, it could still be that \textit{DisclosedToOutsiders} applied.

\textbf{SCQ3}: \textit{Is there another precedent, P2, which can ground an instantiation of the switching point scheme to give an argument that the factor does not (does) apply?}. It may be that some additional information is needed to say which precedent should apply. Suppose there were a case with 200 disclosures where \textit{DisclosedToOutsiders} was held not to apply. Further examination of this case and \textit{National Rejectors} would be needed to explain this different treatment, and the explanation applied to \textit{Leaky}. 
\noindent

\subsection{Trade Off Scheme}

The next scheme concerns trade-offs between two dimensions, as described in \cite{bench2021using}. For example in the US Fourth Amendment domain there is a trade-off between being able to enforce the law and respect for privacy \cite{bench2010using}. The factor involves balancing these two concerns and is something like ``Sufficient respect for privacy while enabling enforcement''. The idea in \cite{bench2021using} is that a line, e.g $a.D_1 + b.D_2 + c = 0$, can be fitted to the precedents\footnote{Of course more complicated curves can be used, but a straight line is the simplest.}, separating the pro-plaintiff and pro-defendant regions of the case space. In the equation $a$ and $b$ are the coefficients of the variables $D_1$ and $D_2$, representing the values on the two dimensions. This determines the gradient of the line which indicates how much more $D_1$ is need to compensate for less $D_2$, and $c$ is a constant showing where the line crosses the axes given these coefficients. 

\smallskip
\indent
\textbf{Trade Off Scheme}

\textit{Precedents Premise}: $P_1 ... P_n$  are precedent cases in which factor $F$ is present.

\textit{Locations Premise}: Precedent $P_i \in \{P_1 .., P_n\}$ have locations $D_{1_i}$ and $D_{2_i}$ for dimensions $D1$ and $D2$,

\textit{Case Premise}: $C_1$ is a case with $L_1$ on dimension $D_1$ and $L_2$ on dimension $D_2$

\textit{Line Premise}: All $a.D_{1_i} + b.D_{2_i} + c \geq$  $0$

\textit{Point Premise}: $a.L_1 + b.L_2 + c \geq (<)$ $0$

\textit{Conclusion}: $F$ applies (does not apply) to $C_1$

\smallskip

\textbf{Example6}
Suppose we have two precedents to which \textit{CompetitiveAdvantage} applied. In one (\textit{Slow}) 8 months  and \$500,000 was saved, and in the other (\textit{Fast}) 24 months were saved, but only \$100,000. A line passing  through these two points is $4(money) + 1(time) - 28 = 0$. Suppose we have a new case, \textit{Useful}, in which \$300,000 and 20 months were saved. We can form the argument

\smallskip

\indent

\textit{Slow} and \textit{Fast} are precedent cases in which \textit{CompetitiveAdvantage} is present

\textit{Slow} and \textit{Fast} have locations (5,8) and (1,24) on the money and time dimensions.

\textit{Useful} has location 3 on the money dimension and 18 on the time dimension

For both \textit{Slow} and \textit{Fast} $4(money) + 1(time) - 28 = 0$

12 + 18 - 28 = 2, which is greater than 0

Therefore, \textit{CompetitiveAdvantage} is present in \textit{Useful}

\smallskip

Now suppose we have a precedent,  \textit{Useless}, in which \$300,000 was saved but only 14 months. Now 12 + 14 - 28 = -2, and so we can argue that \textit{CompetitiveAdvantage} is not present in \textit{Useless}.

\noindent

\smallskip

For this scheme we have the following key critical questions;

\indent
\textbf{TCQ1}: Is there a counter example, a precedent, $P_{n+1}$, for which $a.D_{1_{n+1}} + b.D_{2_{n+1}} + c <(\geq)$  $0$?. There might be a precedent which does not fit the line. For example if we had a precedent which saved \$400,000 and 13 months in which \textit{CompetitiveAdvantage} was held to be absent. This would suggest that we need a more complicated curve than the simple line used in the argument, and it is possible \textit{Useful} would fall on  the wrong side.

\textbf{TCQ2}: Can the line be drawn less (more) tightly? If the precedents are not precisely on the line the constant $c$ could be adjusted to raise (lower) the line to allow (disallow) more cases to qualify unless this created a counter example. For example, if we lower the line by reducing the constant $c$ to 26, then \textit{Useless} will fit and \textit{CompetitiveAdvantage} can also be ascribed to this case.

\noindent


\section{Discussion}

The schemes presented here are derived from reasoning with legal cases in which precedents are available to resolve the issues that arise and the ascription of factors. Some of the procedures can be adapted to practical reasoning in less formal circumstances. For example, when choosing a restaurant \cite{atkinson2013distinctive}, we can identify issues such as value for money, convenience and quality of experience.  Each of these will have associated factors: value will depend on cost and quality, convenience on distance and speed of service, and the quality of experience on the nature of the cuisine and the noise level and so on. Whether these dimensions will give reasons to choose or reject a venue will be a matter of judgement: a restaurant may not be expensive enough that that is a reason to avoid it, nor cheap enough that it is a reason to select it. Thus far, the problem is similar to the legal situation, in that it can be decomposed into the same elements. Some schemes may also be applicable: once the issues have been resolved, we can probably apply a rule: for example that the venue is at least satisfactory on each issue. Identifying reasons for and against also involves the move from points on relevant dimensions to reasons for or against, and this to could make use of the schemes in Section~\ref{ascription}. This big difference lies in the way competing arguments are resolved. In law it is desirable that like cases are treated in a like manner, but this is not so with decisions like choosing a restaurant: in fact variety may be a reason for acting differently from the last time. In such discussions it is important that the preferences accepted by the group be established\footnote{This point applies more generally within deliberation dialogues, as demonstrated through the model provided in \cite{DBLP:conf/sgai/KirchevAB19}.}, but this has to be done other than by appeal to precedent. This need to establish preferences other than by using precedents limits the applicability of these legal schemes to practical reasoning tasks.

Classification tasks are, however, a different matter. The style of explanation given by the above schemes have been used  in \cite{prakken2022top} to explain the outputs from machine learning systems for a variety of non legal domains: customer churn, poisonous mushrooms and university admissions. In such domains it is possible to treat the known cases as precedents, and to explain a classification in terms of similarity of features. If I know one red mushroom with white spots is poisonous, this may well explain why I think a new red mushroom with white spots is also poisonous. Note that in such a domain there are no preferences: the relationship between features and classification is assumed to be a matter of causal \textit{fact}, not personal \textit{choice}, as in the case of practical reasoning tasks. The \textit{direction of fit} \cite{searle2003rationality} is crucial: for classification we must fit our beliefs to the world, whereas in practical reasoning we attempt to make the world fit our desires.

Thus it may be possible to use these schemes in domains outside law, but this will require that the reasoning does not require preferences, as with classification, or that these preferences can be established by some means.


\section{Concluding Remarks}


In this paper we have modelled reasoning with legal cases as a set of argument schemes, with different schemes relevant to the different stages of the process. These scheme will in particular support the presentation of justifications for and explanations of the reasoning in such cases. By presenting instantiations of the schemes and allowing the user to pose critical questions it is possible to tailor the explanations to the needs of particular user, and provide contrastive explanations \cite{miller2019explanation} by explaining why things do not hold as well as why they do.

\bibliographystyle{plain} 
\bibliography{arg22-long}

\end{document}